\documentclass[conference]{IEEEtran}

\usepackage[utf8]{inputenc}
\usepackage{graphicx}
\usepackage{hyperref}
\usepackage{amsmath}
\usepackage{cite}
\usepackage{caption}
\usepackage{booktabs}

\title{The Hidden Costs of AI: A Review of Energy, E-Waste, and Inequality in Model Development}
\author{
\IEEEauthorblockN{Jenis Winsta}
\IEEEauthorblockA{
Independent Researcher\\
jeniswinsta@gmail.com}
}

\begin{document}
\maketitle

\begin{abstract}
Artificial intelligence (AI) has made remarkable progress in recent years, yet its rapid expansion brings overlooked environmental and ethical challenges. This review explores four critical areas where AI’s impact extends beyond performance: energy consumption, electronic waste (e-waste), inequality in compute access, and the hidden energy burden of cybersecurity systems. Drawing from recent studies and institutional reports, the paper highlights systemic issues such as high emissions from model training, rising hardware turnover, global infrastructure disparities, and the energy demands of securing AI. By connecting these concerns, the review contributes to Responsible AI discourse by identifying key research gaps and advocating for sustainable, transparent, and equitable development practices. Ultimately, it argues that AI’s progress must align with ethical responsibility and environmental stewardship to ensure a more inclusive and sustainable technological future.
\end{abstract}

\begin{IEEEkeywords}
Artificial Intelligence, Responsible AI, Green AI, Compute Inequality, E-waste, Sustainability, Energy Efficiency
\end{IEEEkeywords}

\section{Introduction}
Artificial intelligence (AI) has made rapid advances in recent years, with models like GPT-4, Gemini, and LLaMA-3 demonstrating capabilities once thought to be uniquely human. From powering autonomous systems and healthcare diagnostics to generating art and code, AI is reshaping nearly every sector of the global economy.

However, this technological progress comes with a set of hidden environmental and ethical costs. While most attention remains focused on performance metrics and innovation, less scrutiny has been given to the infrastructure supporting large-scale AI systems. Training and deploying these models requires enormous computational power, contributing significantly to carbon emissions, electronic waste, and unequal access to computing resources.

This review examines four underexplored dimensions of AI’s environmental and ethical footprint. First, it investigates the energy demands of model training and deployment, highlighting the carbon intensity of current development practices. Second, it considers AI’s contribution to electronic waste, especially through high hardware turnover and data center infrastructure. Third, it explores the global imbalance in access to compute, which concentrates AI research within a few elite institutions and regions. Finally, it addresses how cybersecurity—essential to safeguarding AI systems—introduces its own energy overheads and infrastructure disparities.

By synthesizing findings from recent research and reports, this paper contributes to the broader conversation on Responsible AI. It identifies key sustainability challenges, outlines gaps in current practices, and emphasizes the need for carbon-aware design, equitable infrastructure, and transparent reporting.

The sections that follow address each of these four themes in turn, concluding with an overview of emerging solutions, research gaps, and future directions. In doing so, the paper advocates for a more ethical and environmentally sustainable approach to AI development—one that matches technological power with responsibility.


\section{Energy Consumption in AI: Emissions, Efficiency, and Accountability}
The recent surge in large-scale artificial intelligence models has led to transformative breakthroughs across fields, ranging from language processing to image generation. However, these gains come with significant hidden costs. One of the most notable is the vast amount of energy required to train, fine-tune, and deploy these models. As AI continues to scale, its environmental footprint also increases, raising urgent concerns about both sustainability and equity.

One of the earliest major analyses of this issue was conducted by Strubell et al. ~\cite{strubell2019energy}, who revealed that training a single large NLP model using neural architecture search emitted over 626,000 pounds of CO$_2$. This is roughly equivalent to the lifetime emissions of five American cars. Their study also highlighted that experimentation, hyperparameter tuning, and retraining contribute significantly to energy consumption beyond initial model training. For example, while training a single Transformer model without tuning emitted 1,438 pounds of CO$_2$, the full experimentation pipeline increased emissions to 78,468 pounds.

\begin{figure}[htbp]
\centering
\includegraphics[width=\linewidth]{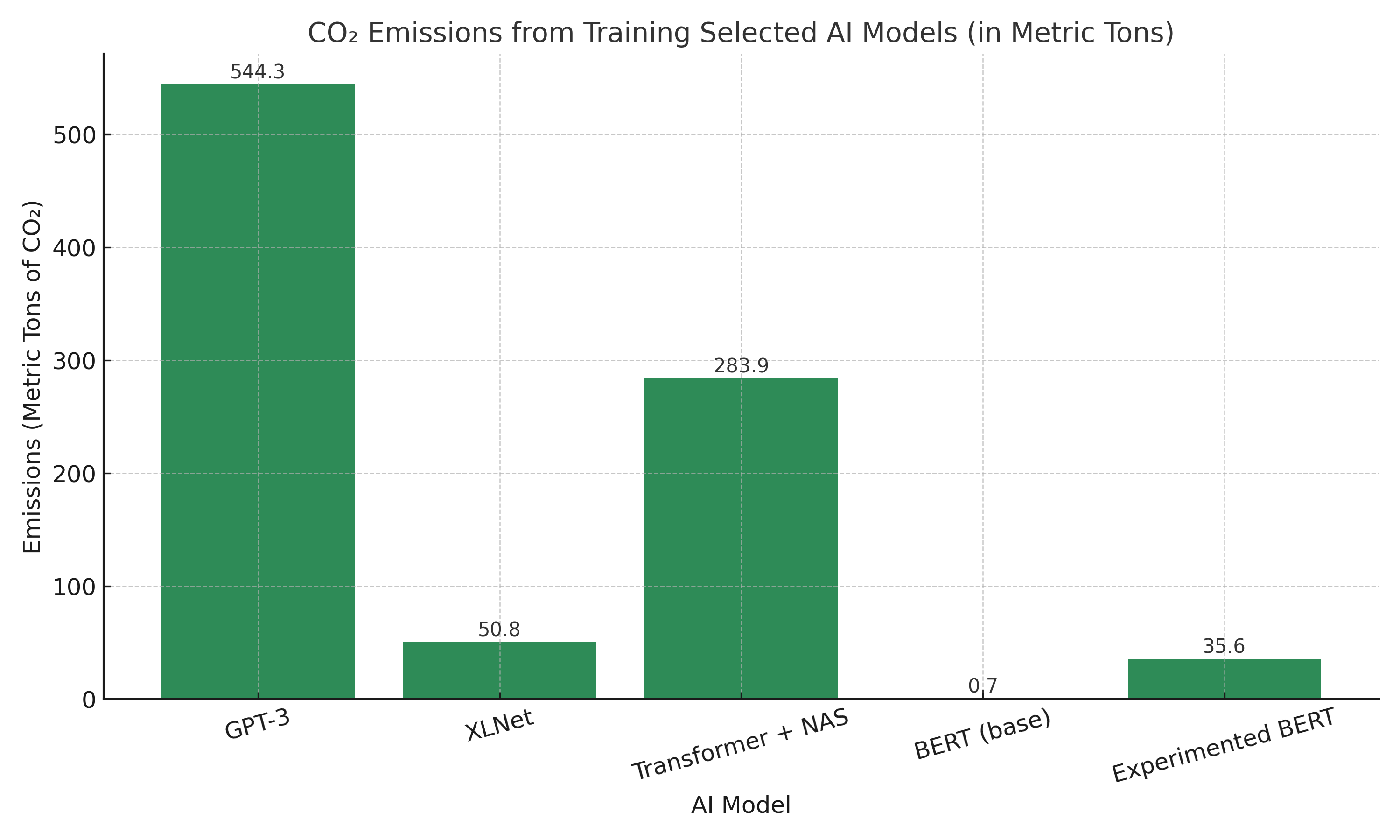}
\caption{CO$_2$ emissions from selected AI models, illustrating the environmental cost of large-scale experimentation. (Adapted from~\cite{strubell2019energy,patterson2021carbon})}
\end{figure}

To visualize these differences in emissions, Figure 1 presents a comparison of CO$_2$ output from several widely studied AI models. It illustrates the sharp increase in emissions as the complexity of the model and experimentation scale up.

Complementing this, Schwartz et al. ~\cite{schwartz2020greenai} introduced the concept of "Green AI", calling for a shift from performance-only benchmarks to energy-efficient and cost-efficient development. They noted that AI’s compute requirements increased by 300,000× between 2012 and 2018, with minimal transparency around energy or emissions. Models like XLNet, trained on 512 TPUs over 2.5 days, cost up to \$250,000, making them both environmentally costly and financially inaccessible to many researchers.

Patterson et al. ~\cite{patterson2021carbon} further demonstrated how energy consumption and CO$_2$ emissions vary dramatically depending on model architecture, hardware, datacenter efficiency, and geographic location. Training GPT-3, for instance, required 1,287 megawatt-hours (MWh) of energy and emitted 552 tons of CO$_2$e equivalent, an amount comparable to multiple transcontinental flights. They advocated for adopting sparse models (e.g., Switch Transformer), which activate fewer parameters per token, reducing compute needs by up to 90\%. Their analysis also showed that choosing energy-efficient chips (for example, Google's TPUs) and greener datacenter locations (for example, Iowa vs Taiwan) could reduce emissions by factors of 10 to 100.

The Stanford Center for Research on Foundation Models (CRFM) expanded this conversation by framing energy consumption as a systemic issue. In their 2021 report, they argued that most emission estimates are incomplete because they focus solely on training and exclude fine-tuning, inference, and deployment. These omitted stages often account for the majority of real-world energy use ~\cite{brundage2020trustworthy}. The report called for a multifaceted approach to reduce AI’s carbon burden, including investments in green infrastructure, greater model transparency, and more equitable access to public computing resources. These steps, they noted, are critical to preventing a further widening of the resource gap between elite AI labs and academic institutions.

In support of this view, Brundage et al. ~\cite{brundage2020trustworthy} emphasized that verifiability in AI development must include transparent and precise measurement of compute usage. They proposed a framework that incorporates secure hardware, institutional coalitions, and third-party auditors to report and regulate energy consumption. Their research highlighted that “high-precision compute tracking” is essential to make credible environmental claims. It also revealed that academic researchers often lack the infrastructure and funding needed to participate in these high-emission AI development environments.

Together, these studies show that AI’s environmental impact is not merely a byproduct but rather a result of deliberate design choices. It is shaped by the ways in which models are built, trained, and deployed. As training emissions continue to increase alongside global deployment, energy use must be considered a core metric in AI evaluation, on equal footing with accuracy and speed. Without systemic changes in reporting practices, accountability measures, and infrastructure support, AI development risks becoming increasingly unsustainable and exclusive.

\section{AI and E-Waste}
As artificial intelligence (AI) systems scale, so too does their reliance on increasingly complex and resource-intensive hardware—leading to a substantial but often overlooked side effect: the growing burden of electronic waste (e-waste). While the environmental discussion around AI typically focuses on emissions from model training, its hardware lifecycle—spanning chips, GPUs, data center infrastructure, and consumer AI devices—contributes significantly to global material and toxic waste streams.
According to the Global E-waste Monitor 2020, the world generated 53.6 million metric tons (Mt) of e-waste in 2019, a figure projected to rise to 74.7 Mt by 2030. Only 17.4\% of this waste is formally collected and recycled, while the remaining 82.6\% is either landfilled, incinerated, or exported—often illegally—to countries lacking proper disposal infrastructure ~\cite{ewaste2020}. Much of this e-waste contains hazardous substances like mercury, brominated flame retardants (BFRs), and chlorofluorocarbons (CFCs), which pose severe environmental and health risks when not properly treated. As AI continues to fuel demand for specialized hardware—from GPUs and TPUs to edge devices and robotics—its role in accelerating hardware turnover is increasingly concerning.
\begin{figure}[htbp]
\centering
\includegraphics[width=\linewidth]{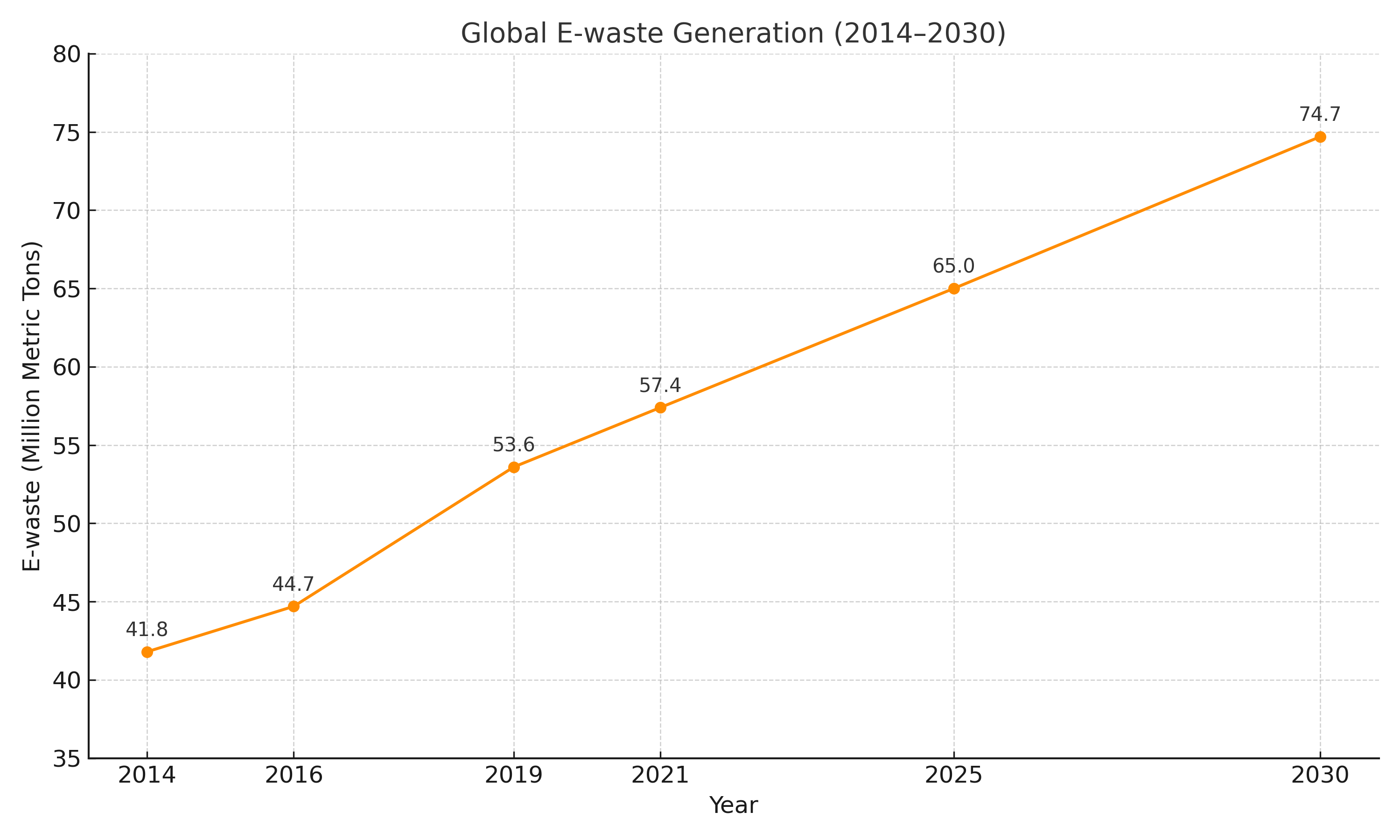}
\caption{Global e-waste generation from 2014 to projected levels in 2030. E-waste is expected to increase from 41.8 million metric tons in 2014 to 74.7 Mt by 2030, driven by rapid hardware turnover and digital infrastructure growth.
Source: Global E-waste Monitor 2020 ~\cite{ewaste2020}.}
\end{figure}

The situation is compounded by the energy demands of AI-supporting infrastructure, particularly data centers, which are essential for training, inference, and storage. As reported in Greenpeace’s Clicking Clean (2017), the IT sector already consumes over 7\% of global electricity, a number projected to rise to 12\% or more ~\cite{cook2017clicking}. AI workloads contribute directly to this demand, especially through cloud-based models and inference services hosted in large-scale data centers. While companies like Apple and Google have taken steps toward using renewable energy, others—including major AI cloud providers like AWS and Tencent—lack transparency about their energy sourcing and carbon footprint. Moreover, many of these data centers are located in regions with fossil-fuel-heavy grids (e.g., Virginia in the U.S.), limiting the real-world impact of renewable energy commitments.

Even as infrastructure becomes more energy-efficient, the Jevons paradox remains relevant—gains in efficiency are frequently offset by a corresponding rise in usage. The widespread deployment of AI systems across streaming services, surveillance networks, smart devices, and recommendation engines has created a continuous and growing demand for computational resources and data storage. This persistent usage accelerates hardware wear and shortens replacement cycles. As a result, electronic waste is emerging not as a byproduct, but as a systemic and escalating consequence of AI-driven growth.

Measuring and regulating this impact remains a significant challenge. The IEA’s EDNA report (2022) points out that commonly used metrics such as Power Usage Effectiveness (PUE), which compare energy used by IT equipment to the total facility load, fail to account for true compute efficiency or energy productivity ~\cite{iea2022metrics}. A data center may score well on PUE but still waste considerable energy if its servers remain underutilized or idle. Research shows that idle servers can consume between 30\% and 70\% of their peak power, further highlighting the inefficiencies embedded in large-scale AI infrastructure. The report calls for globally harmonized and policy-relevant metrics that measure compute, storage, and network output relative to total energy use. This is a critical step toward understanding and mitigating the environmental cost of AI hardware.

Collectively, these findings indicate that AI’s contribution to e-waste extends well beyond consumer devices. It is deeply rooted in the infrastructure that supports modern machine learning. Addressing this challenge requires a broader definition of Responsible AI—one that encompasses hardware lifecycle management, infrastructure transparency, and sustainable procurement and disposal practices. Without meaningful structural reforms in the way AI systems are designed, powered, and retired, the field risks intensifying global inequality, environmental degradation, and resource scarcity under the banner of digital progress.

\section{Inequality in Compute Access}
The development of frontier AI models—such as GPT-4, Gemini, and LLaMA-3—requires immense computational resources that are accessible only to a limited group of elite institutions and large technology companies. Organizations like OpenAI, Google DeepMind, Meta, and Anthropic dominate this space due to their proprietary infrastructure, expansive data pipelines, and unrestricted access to high-performance computing (HPC). This concentration of capability has given rise to the so-called “compute divide,” which refers to the growing gap between those who can afford to develop cutting-edge AI and those who cannot ~\cite{ahmed2020dedemocratization}.

In contrast, academic institutions and low-resource countries face substantial barriers to participating in foundational AI research. According to Brundage et al. ~\cite{brundage2020trustworthy}, the lack of affordable and scalable computing infrastructure forces many researchers to limit their work or transition to industry settings, where access to compute is easier but research agendas may be commercially driven. Even when researchers possess the required expertise, they often lack the infrastructure, funding, and geopolitical support necessary for meaningful participation. The issue is not solely technical—it is systemic and structural.

\begin{figure}[htbp]
\centering
\includegraphics[width=\linewidth]{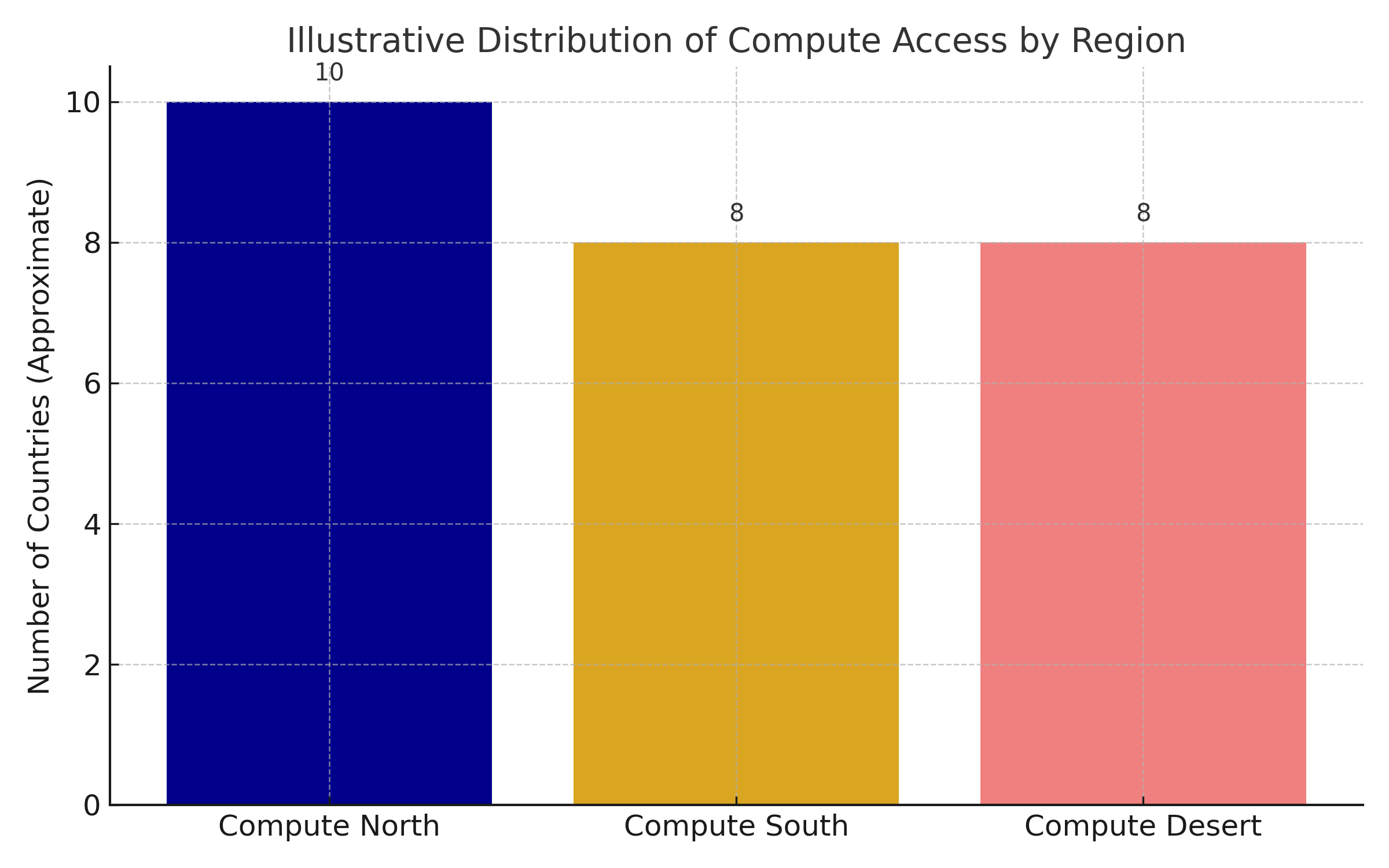}
\caption{Illustrative comparison of countries by access to advanced AI compute infrastructure. “Compute North” includes countries with abundant cloud GPU infrastructure and AI investment; “Compute South” refers to countries with moderate access; “Compute Desert” indicates regions largely excluded from frontier model development due to lack of infrastructure.
 Source: Adapted from Lehdonvirta et al. ~\cite{lehdonvirta2024compute}, Ahmed \& Wahed ~\cite{ahmed2020dedemocratization}, and analysis by the author.}
\end{figure}

Meanwhile, scholars such as Timnit Gebru and Abeba Birhane have framed this divide as a form of “AI colonialism,” where the Global South supplies data and human labor, but the benefits of AI innovation remain concentrated in the Global North ~\cite{mohamed2022decolonial}. Many developing countries participate in data labeling or serve as test populations, yet lack the infrastructure to build, adapt, or govern the tools they help create. As Lehdonvirta et al. ~\cite{lehdonvirta2024compute} show, only a handful of countries host advanced GPU regions, leaving large portions of the world—labeled “Compute Deserts”—excluded from frontier AI development.

This disparity is further exacerbated by the closed nature of most state-of-the-art models. For example, GPT-4 and Gemini remain proprietary and inaccessible. Although open-source alternatives like Meta’s LLaMA and Mistral offer some relief, they still require significant computing resources beyond the reach of many researchers. Cloud credit programs from providers such as AWS, Google, and HuggingFace attempt to bridge this gap but are often short-term, inconsistent, or biased toward already connected regions ~\cite{brundage2020trustworthy}.

In response, efforts toward decentralized and democratized compute are beginning to emerge. Initiatives led by organizations such as HuggingFace, Together AI, and the DAIR Institute (co-founded by Gebru) promote distributed training infrastructure, particularly targeting underrepresented regions ~\cite{dair2024decentralized}. Additionally, Mozilla and UNESCO have called for inclusive AI governance frameworks that address global disparities in access and opportunity ~\cite{mozilla2020trustworthy}.

Ultimately, discussions about compute access must prioritize fairness, transparency, and equitable benefit-sharing. If only a small number of entities can build, train, and deploy powerful AI models, the rewards of AI will remain unequally distributed. Creating a more inclusive and responsible AI future requires that more people—not fewer—have the tools and opportunities to participate in shaping its development.

\section{Cybersecurity and Hidden Energy Cost}
As artificial intelligence (AI) systems expand across industries, securing them is essential—but the energy and environmental impacts of cybersecurity often go unnoticed. Activities such as encryption, anomaly detection, access control, and logging run continuously in the background, consuming substantial computational power, especially at scale. A comparative study found that standard protection systems like antivirus, firewall monitoring, and encrypted network flows incur notable energy overheads, particularly in latency-sensitive deployments ~\cite{smith2023eco}.

Data centers that support large-scale AI models face added complexity. Operating under "zero-trust" frameworks, they require encrypted and logged traffic, which can raise energy use by up to 30\%, depending on workload and cryptographic methods ~\cite{iea2022metrics}. Additionally, compliance systems and redundant security logging increase storage demands, cooling needs, and network load—especially in sensitive sectors like finance and defense.

AI-based cybersecurity systems introduce dual energy burdens. They reduce manual oversight but demand constant updates, inference, and deployment across 24/7 IoT infrastructures—often lacking energy efficiency ~\cite{smith2023eco}. Compliance with regulations such as data sovereignty, audit trails, and the EU AI Act further amplifies indirect energy use through encrypted storage and geo-redundant backups ~\cite{brundage2020trustworthy}.

Addressing these concerns requires energy-conscious cybersecurity design. Lightweight cryptographic protocols, energy-aware intrusion detection, and optimized AI security agents are essential steps ~\cite{brundage2020trustworthy}. Efforts like the Green Software Foundation and Carbon-Aware SDKs are beginning to address this intersection, but adoption remains limited ~\cite{mozilla2020trustworthy}.

Ultimately, cybersecurity must be treated as a key factor in AI’s energy lifecycle. Secure-by-design and green-by-default approaches are needed to balance protection with sustainability.

\section{Solutions, Gaps, and Future Directions}
A promising response to AI’s growing environmental impact is the adoption of Green AI principles. Instead of optimizing solely for performance, Green AI emphasizes computational efficiency, reduced carbon emissions, and broader accessibility as fundamental design goals. Techniques such as model compression, transfer learning, and early-exit strategies enable models to maintain strong performance with significantly lower energy demands. At the same time, the rise of edge AI—which runs inference locally on devices—helps reduce reliance on energy-intensive cloud infrastructure. Tools like CodeCarbon and MLCO2 offer developers insight into emissions during training and deployment, supporting better energy optimization ~\cite{schwartz2020greenai,patterson2021carbon}.

Achieving long-term sustainability will also require embedding these principles into AI education and practice. Curricula in computer science and data science must train students to weigh not just accuracy, but also energy use, carbon footprint, and hardware efficiency. Early exposure to these values can shift the research culture from "bigger is better" to "smarter is better", fostering critical thinking about scale and resource trade-offs ~\cite{schwartz2020greenai, mozilla2020trustworthy}.

In parallel, regulatory and policy efforts are gaining momentum. Proposed measures include mandating emissions disclosures, publishing compute usage and hardware benchmarks, and incentivizing the development of low-carbon AI models. Organizations such as UNESCO, the Mozilla Foundation~\cite{mozilla2020trustworthy}, and the DAIR Institute~\cite{dair2024about} advocate for governance frameworks that prioritize sustainability and equitable access. Emerging legislation such as the EU AI Act presents an opportunity to embed environmental accountability into model certification and reporting processes~\cite{dair2024about}.

However, several critical gaps persist. Most AI research papers—even from leading institutions—still fail to report key details like compute hours, hardware configurations, or training emissions ~\cite{ahmed2020dedemocratization,brundage2020trustworthy}. There are also no binding global regulations on AI-related e-waste, despite rising hardware turnover and the widespread use of GPUs and specialized accelerators ~\cite{ewaste2020}. Perhaps most concerning is the global compute imbalance, which restricts meaningful participation in cutting-edge AI development to a small set of well-resourced institutions and nations ~\cite{ahmed2020dedemocratization,lehdonvirta2024compute}.

Bridging these gaps will require collaborative efforts across academia, industry, and governments. Publicly funded compute infrastructure, transparent emissions tracking, and open-source model initiatives can play a key role in democratizing access ~\cite{odsc2024huggingface}. Research funding agencies should prioritize sustainability-driven grants, while cloud providers must ensure equitable distribution of compute resources in underserved regions ~\cite{brundage2020trustworthy}. Policy frameworks should also address green procurement, ethical hardware recycling, and global AI standards that integrate sustainability and social equity alongside performance and safety ~\cite{iea2022metrics,mozilla2020trustworthy}.

In the end, AI’s future must be not just intelligent and powerful, but also ethical, inclusive, and environmentally responsible. Redefining progress to include sustainability metrics will help ensure that the benefits of innovation are shared more equitably—and that its environmental and social costs are not silently passed on to vulnerable communities or future generations ~\cite{schwartz2020greenai}.

\section{Conclusion}
Artificial Intelligence is among the most transformative technologies of the 21st century, driving breakthroughs in language modeling, healthcare, and industrial automation. Yet, behind these advancements lie underexamined environmental and ethical costs. This review has highlighted AI’s hidden footprint—its rising energy consumption, growing contribution to global e-waste, inequality in compute access, and the overlooked energy burden of cybersecurity infrastructure.

These concerns are real and immediate. Training state-of-the-art models today can emit hundreds of tons of CO$_2$, while the hardware supporting these systems accelerates e-waste generation. At the same time, access to the compute resources needed to build frontier models remains concentrated in a handful of institutions and nations, raising pressing questions about fairness and inclusion.

Without meaningful reforms, the gap between AI’s creators and the communities affected by it will continue to widen. Accountability must become a foundational element of AI development. This includes emissions transparency, sustainable hardware practices, and regulatory frameworks that align innovation with environmental boundaries.

Still, there are reasons for optimism. Awareness of these issues is growing, and with it, a movement toward Green AI, energy-efficient modeling, and inclusive infrastructure. Researchers, activists, and policymakers are beginning to redefine success in AI—not just through performance, but through sustainability and social impact.

For AI to truly benefit humanity, its development must be grounded in responsibility. By embedding sustainability and ethics at its core, the AI community can shape a future where innovation does not come at the expense of the environment or equity. Recognizing AI’s hidden costs is the first step toward creating systems that are not only powerful and intelligent, but also just and sustainable.

\bibliographystyle{IEEEtran} 
\bibliography{reference}  
\end{document}